\documentclass[sigconf]{acmart}
\usepackage{amsmath,graphicx}
\usepackage{multirow}
\usepackage{makecell}
\usepackage{amsfonts}
\usepackage{bm}
\usepackage{tabularx}

\usepackage{bbding}
\AtBeginDocument{%
  \providecommand\BibTeX{{%
    \normalfont B\kern-0.5em{\scshape i\kern-0.25em b}\kern-0.8em\TeX}}}

\begin{document}

%%
%% The "title" command has an optional parameter,
%% allowing the author to define a "short title" to be used in page headers.
\title{Uncertainty-aware Evidential Fusion-based Learning for Semi-supervised Medical Image Segmentation}

%%
%% The "author" command and its associated commands are used to define
%% the authors and their affiliations.
%% Of note is the shared affiliation of the first two authors, and the
%% "authornote" and "authornotemark" commands
%% used to denote shared contribution to the research.
 \author{Yuanpeng He}
%\authornotemark[1]
\email{heyuanpeng@stu.pku.edu.cn}
\affiliation{%
	\institution{Key Laboratory of High Confidence Software Technologies (Peking University), Ministry of Education}
	%\streetaddress{P.O. Box 1212}
	\city{Beijing}
	%\state{Ohio}
	\country{China}
	\postcode{100871}
}

\author{Lijiang Li}
%\authornotemark[2]
\email{mc35305@umac.mo}
\affiliation{%
	\institution{Department of Computer and Information Science, University of Macau}
	\city{Macau}
	\country{China}
	\postcode{999078}
}
\begin{abstract}
  Although the existing uncertainty-based semi-supervised medical segmentation methods have achieved excellent performance, they usually only consider a single uncertainty evaluation, which often fails to solve the problem related to credibility completely. Therefore, based on the framework of evidential deep learning, this paper integrates the evidential predictive results in the cross-region of mixed and original samples to reallocate the confidence degree and uncertainty measure of each voxel, which is realized by emphasizing uncertain information of probability assignments fusion rule of traditional evidence theory. Furthermore, we design a voxel-level asymptotic learning strategy by introducing information entropy to combine with the fused uncertainty measure to estimate voxel prediction more precisely. The model will gradually pay attention to the prediction results with high uncertainty in the learning process, to learn the features that are difficult to master. The experimental results on LA, Pancreas-CT, ACDC and TBAD datasets demonstrate the superior performance of our proposed method in comparison with the existing state of the arts. 
\end{abstract}
  % Moreover, to the best of our knowledge, it is the first time to introduce evidential fusion-based uncertainty measure and uncertainty-based voxel-wise asymptotic learning into the semi-supervised medical image segmentation task.
%%
%% The code below is generated by the tool at http://dl.acm.org/ccs.cfm.
%% Please copy and paste the code instead of the example below.
%%
% \begin{CCSXML}
% <ccs2012>
%  <concept>
%   <concept_id>00000000.0000000.0000000</concept_id>
%   <concept_desc>Do Not Use This Code, Generate the Correct Terms for Your Paper</concept_desc>
%   <concept_significance>500</concept_significance>
%  </concept>
%  <concept>
%   <concept_id>00000000.00000000.00000000</concept_id>
%   <concept_desc>Do Not Use This Code, Generate the Correct Terms for Your Paper</concept_desc>
%   <concept_significance>300</concept_significance>
%  </concept>
%  <concept>
%   <concept_id>00000000.00000000.00000000</concept_id>
%   <concept_desc>Do Not Use This Code, Generate the Correct Terms for Your Paper</concept_desc>
%   <concept_significance>100</concept_significance>
%  </concept>
%  <concept>
%   <concept_id>00000000.00000000.00000000</concept_id>
%   <concept_desc>Do Not Use This Code, Generate the Correct Terms for Your Paper</concept_desc>
%   <concept_significance>100</concept_significance>
%  </concept>
% </ccs2012>
% \end{CCSXML}

% \ccsdesc[300]{Do Not Use This Code~Generate the Correct Terms for Your Paper}
% \ccsdesc{Do Not Use This Code~Generate the Correct Terms for Your Paper}
% \ccsdesc[100]{Do Not Use This Code~Generate the Correct Terms for Your Paper}
% %%
% %% Keywords. The author(s) should pick words that accurately describe
% %% the work being presented. Separate the keywords with commas.
% \keywords{Do, Not, Us, This, Code, Put, the, Correct, Terms, for,
%   Your, Paper}
% %% A "teaser" image appears between the author and affiliation
% %% information and the body of the document, and typically spans the
% %% page.
% \begin{teaserfigure}
%   \includegraphics[width=\textwidth]{sampleteaser}
%   \caption{Seattle Mariners at Spring Training, 2010.}
%   \Description{Enjoying the baseball game from the third-base
%   seats. Ichiro Suzuki preparing to bat.}
%   \label{fig:teaser}
% \end{teaserfigure}

% \received{20 February 2007}
% \received[revised]{12 March 2009}
% \received[accepted]{5 June 2009}

%%
%% This command processes the author and affiliation and title
%% information and builds the first part of the formatted document.
\maketitle
Medical image segmentation (MIS) \cite{DBLP:conf/aaai/HuangXL00WLH023,DBLP:conf/cvpr/JiangR00ZNX0023,DBLP:conf/cvpr/ShinKKJEH23} is a critical task in medical image analysis, aiming at identifying and segmenting specific anatomical or pathological structures of interest from image data. The goal of segmentation is to extract accurate and precise information from medical images, which is essential for diagnostic purposes, treatment planning, and disease monitoring. The field of medical imaging encompasses a wide range of image modalities, such as CT scans and MRI, exhibiting distinct distributional differences compared to natural images. Considering hardware or software-related issues during the acquisition process, as well as human physiological phenomena or physical constraints, various types of artifacts may emerge, leading to substantial variance in image quality within medical imaging. Moreover, the annotation of medical images is impeded by the expertise required, making it impractical to address through simple outsourcing. In addition, the complex and time-consuming nature of medical image data acquisition, coupled with limited inter-institutional data exchange, collectively contribute to the scarcity of high-quality medical image data. Acquiring high-quality annotated \cite{najjar2023redefining,ahmed2020artificial} medical datasets is both expensive and challenging. Various techniques based on supervised learning for medical image segmentation have been proposed \cite{dou2020unpaired, li2018h, zhao20213d},
which usually require a large amount of labeled data \cite{DBLP:conf/iccv/WangLG19}. Consequently, within the existing constraints, the exploration of semi-supervised learning approaches for efficient MIS has emerged as a consensus among researchers.

\begin{figure}
    \centering\setlength{\belowcaptionskip}{-0.5cm}
    \includegraphics[scale=0.77]{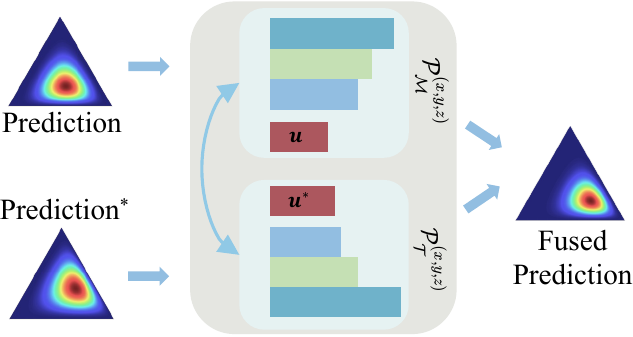}
    \caption{The diagram illustrates the fusion process of the model's predictions for the voxels at the same location in the original samples and the samples after mixed area restoration. It demonstrates the redistribution and adjustment of the voxel's confidence level in belonging to a category after interacting with the overall uncertainty of the prediction.}
    \label{fusion}
\end{figure}

In the field of semi-supervised learning, the limited availability of labeled data \cite{DBLP:conf/nips/BerthelotCGPOR19, DBLP:conf/aaai/Cascante-Bonilla21, DBLP:conf/iclr/RizveDRS21} is overcome by leveraging the abundance of unlabeled data. Through capitalizing on the inherent structure and patterns in the unlabeled data, the model can learn more generalized representations and capture the underlying data distribution more effectively, which allows for enhanced generalization and improved performance when making predictions on new, unseen data. Current semi-supervised medical image segmentation models \cite{DBLP:conf/miccai/LiZH20, DBLP:conf/miccai/YuWLFH19, DBLP:conf/cvpr/BaiCL0023, DBLP:conf/miccai/WuXGCZ21} can be broadly classified into two main approaches. The first approach combines both labeled and unlabeled data during model training, enabling models to learn more generalized semantic features \cite{bai2023bidirectional}. The second approach uses uncertainty estimates to refine and correct the model's segmentation outputs, which involves leveraging the inherent unpredictability and variability in medical image data to enhance segmentation accuracy \cite{wang2022semi,yao2022enhancing,hua2022uncertainty,li2021semantic}. However, relying on uncertainty from a single source can lead to excessive trust in specific data, overlooking other variables or potential biases that might affect the outcomes. Furthermore, if the uncertainty from a single source is overestimated or underestimated, it may misguide decision-making processes based on these uncertainties, thereby impacting the quality and effectiveness of the decisions conducted by models.

To address these issues, we propose a novel strategy for uncertainty estimation different from existing researches \cite{DBLP:conf/miccai/YuWLFH19, DBLP:journals/mia/XiaYYLCYZXYR20,DBLP:journals/corr/abs-2205-10334, DBLP:conf/miccai/LuoLCSCZCWZ21} using improved probability assignments fusion (IPAF, presented in Fig. \ref{fusion}) and voxel-wise asymptotic learning (VWAL). Concerning IPAF, it further strengthens the association between uncertainty and confidence degree of each category for more significant uncertainty indication and better balances important information in the different-sourced prediction. Besides, VWAL is devised to combine information entropy with fused uncertainty measure in evidential deep learning framework to asymptotically guide the model into a fine-grained learning process that gradually focuses on voxels which are difficult to learn and eliminates the potential confirmation bias between labeled and unlabeled data \cite{DBLP:conf/ijcnn/ArazoOAOM20}. The innovative approach aims to provide models with a nuanced understanding of uncertainty to facilitate more fine-grained knowledge mining. 
% To the best of our knowledge, this is the first attempt to incorporate an evidential fusion-based uncertainty measure and uncertainty-based voxel-wise asymptotic learning into semi-supervised medical image segmentation. 
Additionally, the proposed method has achieved state-of-the-art performance on four popular medical benchmark datasets.

The main contributions of this work are summarized as follows:
\begin{itemize}
    \item IPAF provides a comprehensive confidential degree and uncertainty measure interaction among evidential predictive results through effectively combining predictions from cross regions of mixed and original samples.
    \item The strategy of VWAL considers uncertainty of different sources holistically and guides the model to further pay attention to hard-to-learn features by sorting voxel-level optimized uncertainty measure.
    \item The proposed method realizes the state-of-the-art (SOTA) performance on the task of semi-supervised medical image segmentation on four popular benchmark datasets and achieves performance that surpasses methods utilizing 20\% labeled data with only 5\% labeled data on TBAD dataset.
\end{itemize}

The remaining parts of this paper are organized as follows. Section \ref{sec:relatedwork} introduces the related concepts. The details of the proposed method are provided in Section \ref{sec:method} and a series of experiments are conducted in Section \ref{sec:exp}. Finally, conclusions are drawn in Section \ref{sec:con}.
\section{Related Work}
\label{sec:relatedwork}
\subsection{Evidential Deep Learning and Evidence Theory}
% Evidence Theory, also known as Dempster-Shafer Theory (DST) \cite{DBLP:series/sfsc/Dempster08a,DBLP:journals/ijar/Shafer16}, is a mathematical framework for modeling epistemic uncertainty. It provides a flexible tool for combining evidence from different sources and making decisions based on incomplete and uncertain information. Unlike traditional probability theory, which requires precise probabilities for each event, DST works with belief functions that offer a range of probability values, allowing for a more nuanced expression of uncertainty. As deep learning technology continues to evolve, ensuring the reliability of deep learning models and reducing the uncertainty of prediction results have increasingly become hot topics of research. Drawing from the principles of Dempster-Shafer theory of evidence and Subjective Logic \cite{DBLP:books/sp/Josang16}, evidential deep learning (EDL) \cite{amini2020deep,DBLP:conf/nips/SensoyKK18} is introduced to overcome the limitations of softmax-based classifiers discussed. Instead of estimating class probabilities directly, EDL first gathers evidence for each class, and then constructs a Dirichlet distribution of class probabilities based on the evidence obtained. This approach allows for the quantification of predictive uncertainty through subjective logic. Evidence is conceptualized as the degree of support amassed from the data for classifying a sample into a specific class \cite{DBLP:series/sfsc/LiuY08}, serving as a quantifiable measure of class activation intensity.

Evidence Theory, often referred to as Dempster-Shafer Theory (DST) \cite{DBLP:series/sfsc/Dempster08a,DBLP:journals/ijar/Shafer16}, is a robust mathematical structure designed to handle the complexity of epistemic uncertainty. This theory offers a dynamic approach for integrating diverse pieces of evidence and forming judgments when faced with partial and ambiguous data. In contrast to the conventional probability theory, which demands exact probabilities for each scenario, DST employs belief functions \cite{DBLP:conf/ijcai/ZhouQD18}. These functions provide a spectrum of probability estimates, enabling a deeper and more detailed representation of uncertainty. With the ongoing advancements in deep learning technologies, ensuring the dependability of these models and diminishing the unpredictability of their predictions has emerged as a significant area of scholarly interest. Inspired by the foundational concepts of Dempster-Shafer's evidence theory and Subjective Logic \cite{DBLP:books/sp/Josang16}, the concept of evidential deep learning (EDL) \cite{amini2020deep,DBLP:conf/nips/SensoyKK18} has been developed. This methodology seeks to address the inherent shortcomings of softmax-based classifiers. Rather than directly calculating class probabilities, EDL prioritizes the accumulation of class-specific evidence, subsequently utilizing this evidence to formulate a Dirichlet distribution for class probabilities \cite{DBLP:conf/iclr/Xie0ZL23}. This technique facilitates the measurement of predictive uncertainty via subjective logic, where evidence is defined as the accumulated support from the data for categorizing a sample into a particular class \cite{josang2002subjective,koehler2003evidential}. This process acts as a quantifiable indicator of the intensity with which a class is activated.

\subsection{Semi-supervised Medical Image Segmentation}
Semi-supervised medical image segmentation has a wide range of prospects in real-world applications. There have been a large number of related studies proposing various solutions \cite{DBLP:journals/pami/WuZ23,DBLP:conf/cvpr/BasakY23}. Specifically, some researches are developed based on a teacher-student architecture with different strategies. UA-MT \cite{DBLP:conf/miccai/YuWLFH19} proposes a novel uncertainty awareness scheme that enables the student model to use uncertainty information to gradually learn from the meaningful and reliable goals given by the teacher network. SASSNet \cite{DBLP:conf/miccai/LiZH20} designs a deep multi-task network with enforcing geometric shape constraints on the segmentation output. DTC \cite{DBLP:conf/aaai/LuoCSW21} is based on dual-task consistency and the prediction results of unlabeled data and labeled data are used to jointly promote the completion of the segmentation task. URPC \cite{DBLP:conf/miccai/LuoLCSCZCWZ21} introduces the concept of uncertainty correction, which corrects the predictions by calculating the uncertainty of them. MC-Net \cite{DBLP:conf/miccai/WuXGCZ21} proposes a cyclic pseudo-labeling scheme through encouraging consistency between model predictions and true labels, as well as mutual consistency between models. SS-Net \cite{DBLP:conf/miccai/WuWWGC22} achieves more accurate, efficient, and practical segmentation results by simultaneously exploring pixel-level smoothness and inter-class separability. Moreover, BCP \cite{DBLP:conf/cvpr/BaiCL0023} enables the model to better understand the semantic information in images by copying the foreground region of labeled or unlabeled images to the background region of unlabeled or labeled images. 
%It is worth noting that there have been several works that have introduced uncertainty evaluation into the semi-supervised medical segmentation task. They generally use uncertainty to guide the model to learn important features or correct prediction, but ignore the point that the performance of the model not only depends on the mastery of simple features, but also often depends on whether the model can well understand those features which are difficult to be learnt. Obviously, only focusing on important features or modifying predictions does not fundamentally empower the model to learn fully. Therefore, we propose a training strategy based on generalized evidential deep learning, and the uncertainty evaluation scheme based on it can better measure the confidence degree of prediction, which allowing the model to gradually realize the easy-to-difficult learning process and understand each type of feature better. To the best of our knowledge, it is the first time that evidential deep learning $EDL$ and evidential fusion $EF$ are introduced into the semi-supervised medical segmentation task, and our proposed training strategy with generalized $EDL$ and $EF$ achieves the best model segmentation performance so far.
\begin{figure*}
    \centering%\setlength{\belowcaptionskip}{-0.3cm}
    \includegraphics[scale=0.343]{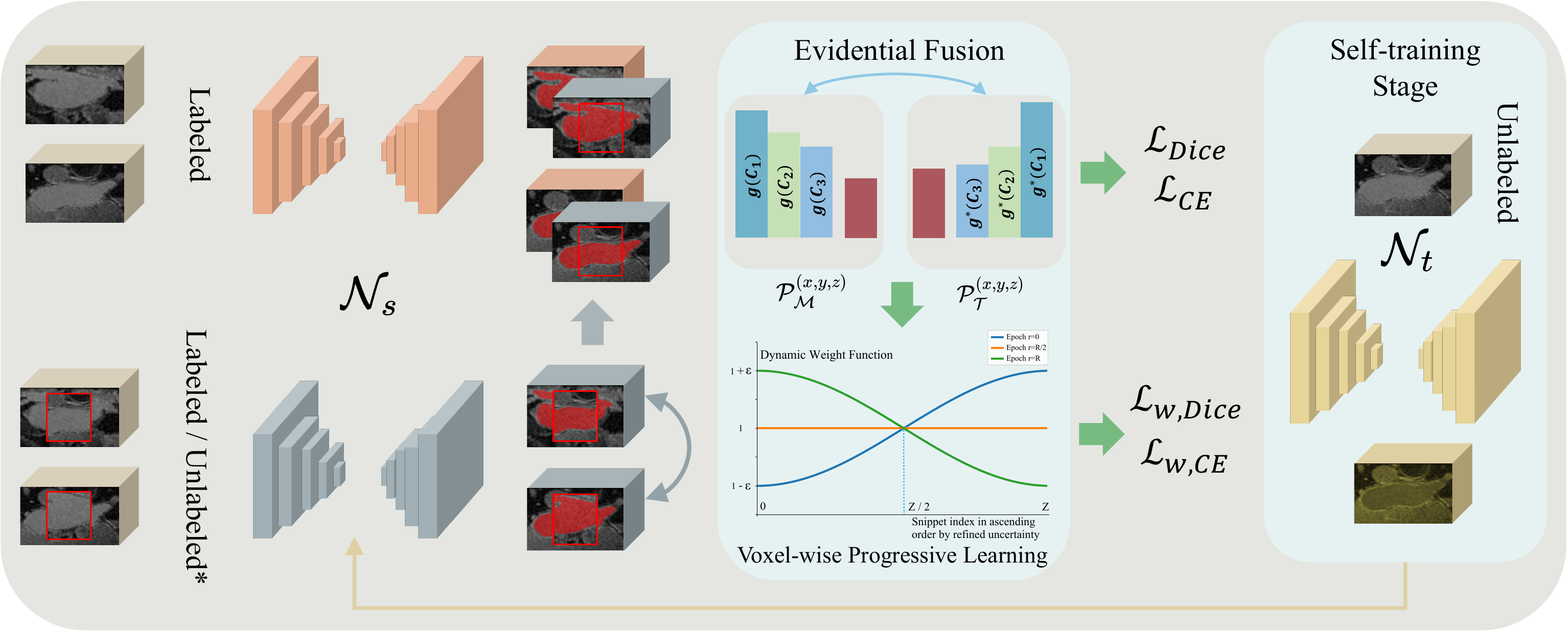}
    \caption{In the evidential fusion-based framework, the pre-training phase involves the student network processing original and mixed data to generate separate predictions. It then restores mixed data to match original samples, employing an enhanced probabilistic strategy for fusing voxel predictions from two sources, considering prediction uncertainty. This process adjusts confidence levels and updates voxel uncertainty, facilitating balanced loss calculation through fused voxel predictions and dynamic weighting. During self-training, the addition of unlabeled data, with labels from the teacher network, is the key change. The teacher network's parameters are updated via the Exponential Moving Average (EMA) method.}
    \label{architecture}
\end{figure*}
\section{Method}
\label{sec:method}
For the task of semi-supervised medical image segmentation, the 3D volume of a medical image can be defined as $X \in \mathbb{R}^{W\times H \times L}$. Semi-supervised medical image segmentation is designed to predict a per-voxel label map $\widetilde{Y} \in \{0,1,...,C-1\}^{W\times H \times L}$, which illustrates the position of targets and background in $X$. Besides, $C$ is the number of classes. Assume there exists a dataset consisting of labeled set $\mathcal{T}^{l} = \{X_{i}^{l}, Y_{i}^{l}\}_{i=1}^{A}$ with $A$ samples and unlabeled set $\mathcal{T}^{u} = \{X_{j}^{u}\}_{j=A+1}^{A+B}$ with $B$ samples $(A \ll B)$. The overall pipeline of the proposed evidential fusion-based framework is shown in Fig. \ref{architecture}. Inspired by the previous works \cite{DBLP:conf/cvpr/BaiCL0023}, we can obtain mixed samples $\mathcal{M}^{l}$ that only utilize labeled samples, as well as mixed samples $\mathcal{M}^{ul}$ that utilize both labeled and unlabeled samples.

\subsection{Generalized Probabilistic Framework}
Evidential deep learning has achieved excellent results in multiple fields. We extend the framework of evidential deep learning to semi-supervised medical segmentation tasks by further improving the uncertainty measurement methods and probability assignments fusion rule of traditional evidence theory. We map the uncertainty measures of evidential deep learning to the multi-objective subsets of traditional evidence theory and fuse reliable information for both mixed and original sample annotation areas during the prediction phase, thereby achieving more robust model predictions with refined uncertainty measures. Moreover, drawing inspiration from curriculum learning \cite{bengio2009curriculum}, we design a novel learning mechanism based on the refined uncertainty measure for the model to gradually focus on difficult-to-learn features.

In detail, let $g^{(x,y,z)}_{\mathcal{M}^{l}}, g^{(x,y,z)}_{\mathcal{T}^{l_1}}, g^{(x,y,z)}_{\mathcal{T}^{l_2}},g^{(x,y,z)}_{\mathcal{M}^{ul}}, g^{(x,y,z)}_{\mathcal{T}^{u}} \in \mathcal{R}^{C}$ denote belief mass from evidential predictive results \cite{DBLP:conf/nips/SensoyKK18,10.5555/3618408.3618708} from classifier of initial network $\mathcal{N}$ in pre-training stage on samples $\mathcal{M}^{l}, \mathcal{T}^{l}$, and student network $\mathcal{N}_s$ in self-training stage on samples $\mathcal{T}^{l}, \mathcal{M}^{ul}, \mathcal{T}^{u}$ for voxel at position $(x,y,z)$ (sample indexes are omitted if there is no confusion). The final trained parameters of the initial network $\mathcal{N}$ will be utilized as the parameters of the teacher and student networks in the beginning of the self-training phase. The basic probability mass assignments for voxels can be defined as $\mathcal{P}^{(x,y,z)} = \{\{g^{(x,y,z)}(C_{n})\}_{n=0}^{N-1}, u^{(x,y,z)}\}$ and $u^{(x,y,z)} = 1 - \sum_{n=0}^{N-1}g^{(x,y,z)}(C_{n})$ denotes the original uncertainty measure. The generalized form of predictive probability assignments can be formulated as:
\begin{equation}
\begin{aligned}
    &\mathcal{P}^{(x,y,z)} = \{\{g^{(x,y,z)}(C_{n})\}_{n=0}^{N-1}, g^{(x,y,z)}(C_N)\},\\ &\qquad\qquad\quad\ C_N = \{C_0,...,C_{N-1}\}
\end{aligned}
\end{equation}
\begin{figure*}
    \centering\setlength{\belowcaptionskip}{-0.3cm}
    \includegraphics[scale=0.3]{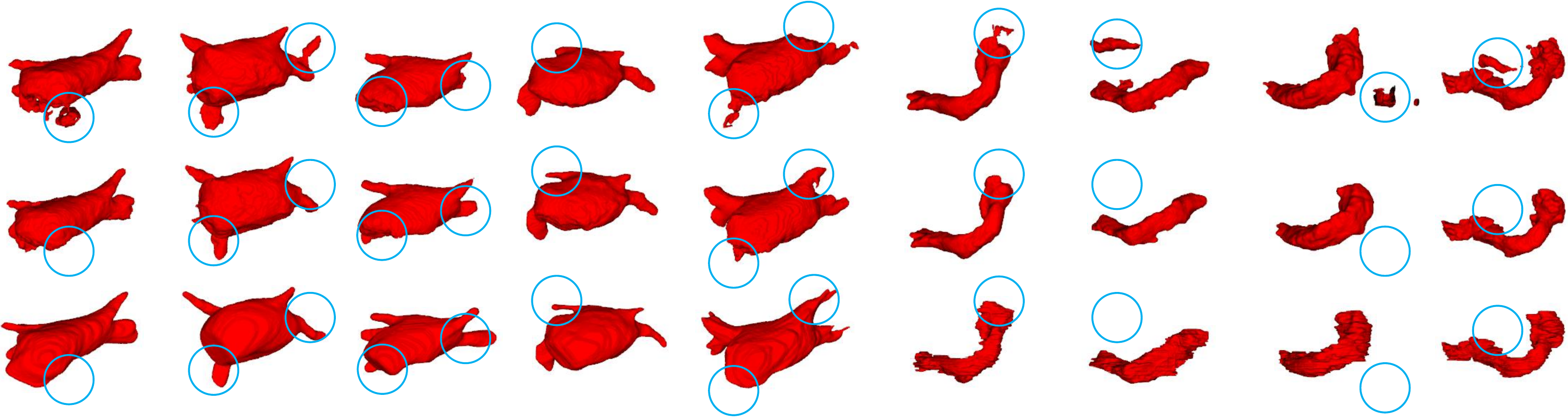}
    \caption{Visualization of experimental results on Left Atrium (LA) and Pancreas-CT dataset. The first line in the visualized image represents the results of the comparative methods (A\&D, BCP). The second line shows the visualized results of the proposed method. The third line indicates the segmentation ground truth corresponding to each image.}
    \label{figure1}
\end{figure*}
\begin{figure}
    \centering\setlength{\belowcaptionskip}{-0.3cm}
    \includegraphics[scale=0.32]{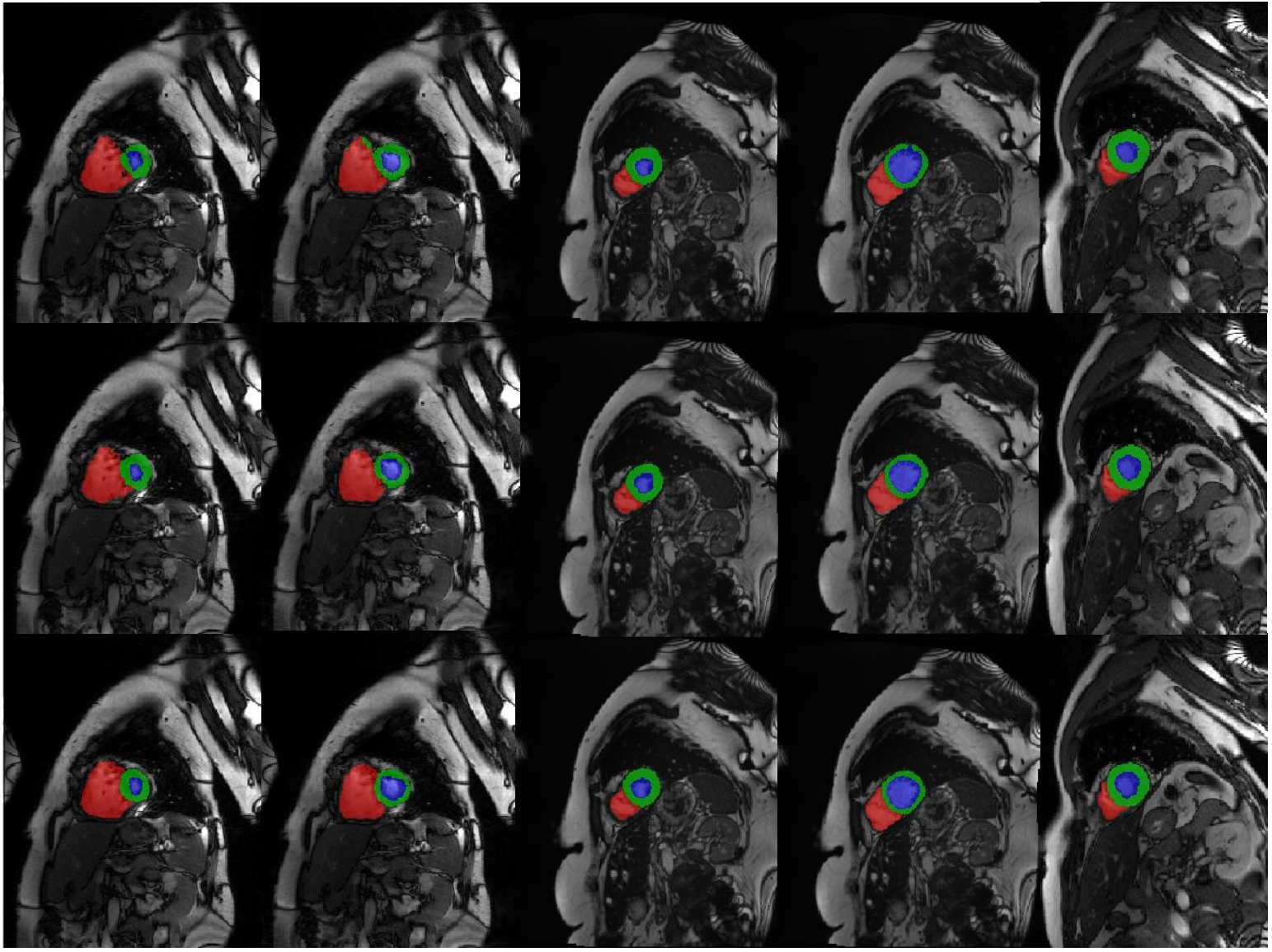}
    \caption{Visualization of experimental results on ACDC dataset. The first line in the visualized image represents the results of the comparative method (A\&D). The second line shows the visualized results of the proposed method. The third line indicates the segmentation ground truth corresponding to each image.}
    \label{figure2}
\end{figure}
\begin{figure}
    \centering\setlength{\belowcaptionskip}{-0.3cm}
    \includegraphics[scale=0.32]{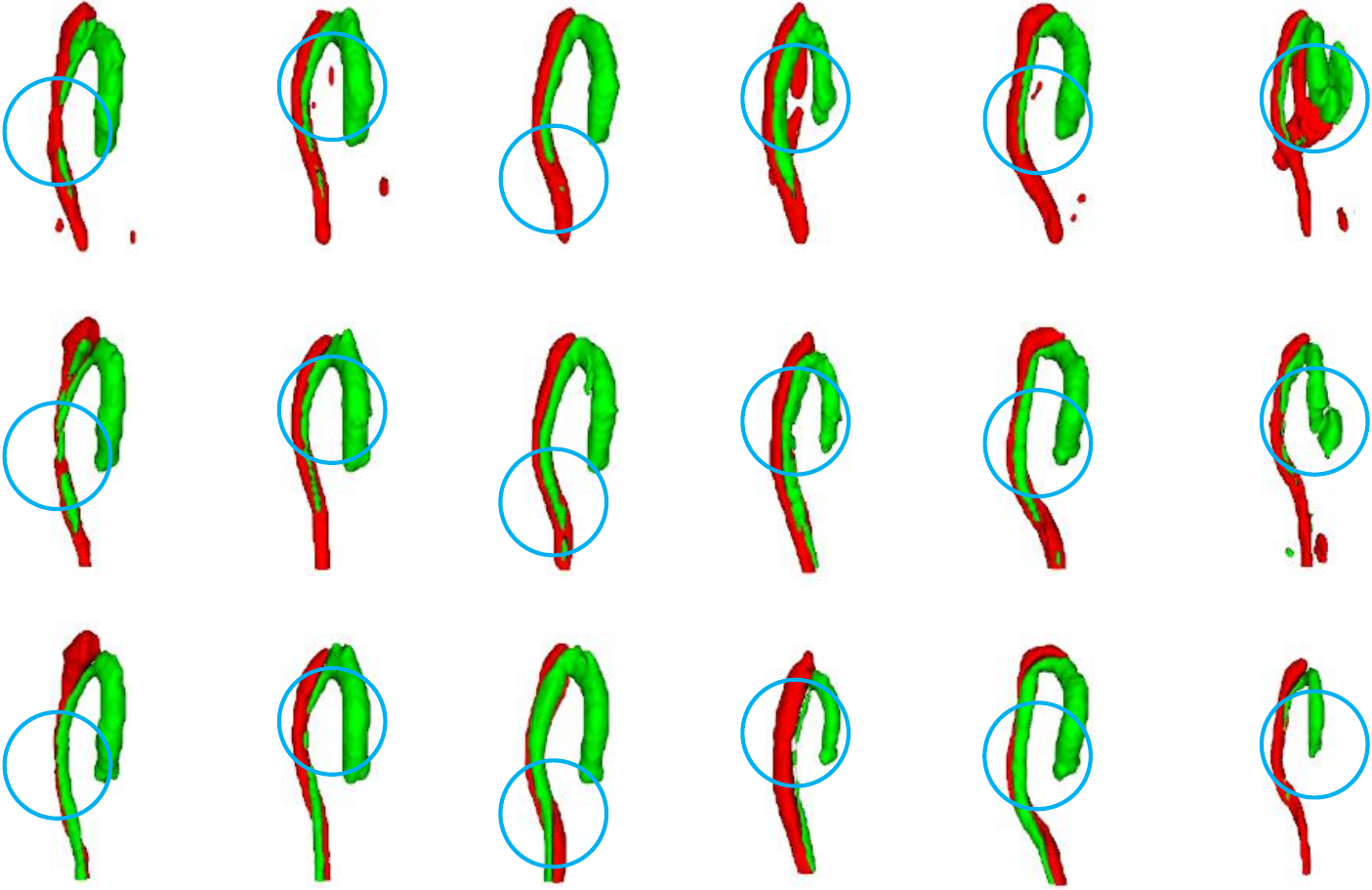}
    \caption{Visualization of experimental results on TBAD dataset. The first line in the visualized image represents the results of the comparative method (UPCoL). The second line shows the visualized results of the proposed method. The third line indicates the segmentation ground truth corresponding to each image.}
    \label{figure3}
\end{figure}

\subsection{Uncertainty-aware Evidential Fusion}
In the previous section, we have mentioned that the method proposed in this article is divided into pre-training and self-training stages. For the pre-training stage, we first feed the samples into the initial network to obtain their predictive results. Due to the uniqueness of the mixed strategy \cite{DBLP:conf/cvpr/BaiCL0023}, we match the original labeled samples with the mixed samples one-to-one, and merge the evidential results of the corresponding areas of the two predictions. More specifically, if there are samples $E$ and $T$, which exchange features in the corresponding area $\mathcal{D}$ to obtain the mixed samples $E'$ and $T'$, after obtaining the evidential predictive results for the original samples $E$, $T$ and the mixed samples $E'$, $T'$, we return the exchanged areas of the mixed samples prediction back to their original positions and fuse them with the model's predictive results for the original samples. Finally, the combined results and prediction of original data are used for loss calculation with the corresponding labels $Y^{l}$ to obtain $\mathcal{L}^{l_1}$, both of which consist of Dice loss and Cross-Entropy loss. We consider the fusion of voxel-level evidential results from two different predictive outputs that come from the same sample and are located at the same position $(x, y, z)$. For restored samples in $\mathcal{M}^{l}$ and original samples in $\mathcal{T}^{l}$, the proposed combination process for voxel at same position can be defined as:
\begin{equation}
    \mathcal{P}^{(x,y,z)}_{fused,l} = \mathcal{P}^{(x,y,z)}_{\mathcal{M}^{l}} \otimes \mathcal{P}^{(x,y,z)}_{\mathcal{T}^{l}}
\end{equation}

Specifically, the strategy of combination can be given as (the superscript $(x, y, z)$ and index of samples are omitted for simplicity):
\begin{equation}
\begin{aligned}
    &g_{fused}(C_n) = g_{\mathcal{M}^{l}}(C_n)g_{\mathcal{T}^{l}}(C_n) + \frac{|C_n|}{|C_N|} \cdot \\&(g_{\mathcal{M}^{l}}(C_n)g_{\mathcal{T}^{l}}(C_N) + g_{\mathcal{T}^{l}}(C_n)g_{\mathcal{M}^{l}}(C_N))
\end{aligned}
\end{equation}
where $n = 0, ..., N-1$, $|C_n| = 1, |C_N| = N$ and $g_{fused}(C_N) = g_{\mathcal{M}^{l}}(C_N) * g_{\mathcal{T}^{l}}(C_N)$. This fusion process evolves from the traditional Dempster's combination rule requiring normalization  \cite{DBLP:series/sfsc/Dempster08a}. The proposed fusion strategy maintains the design for the interaction between the confidence level of each category and the uncertain information, while simultaneously controlling the degree of interaction to avoid the confidence level of each category being dominated by uncertain information. Therefore, the interaction between the confidence associated with each category and the uncertain components is subject to certain restrictions, and we prefer the model to reduce uncertainty through the learning process, rather than directly converting uncertain parts into definite confidence assignments. If the fusion of probabilities follows the original Dempster's combination rule directly, it may lead to the model being overly confident about certain predictions, thereby diminishing the focus on corresponding features, even though learning these features should still be encouraged. Similarly, for samples in the self-training stage, the fusion process they undergo is very similar to that in the pre-training stage. Labeled and unlabeled data undergo the same mixing operation as in the pre-training stage, and after the model makes predictions, the mixed samples are restored and different-sourced belief degrees are fused, then loss calculation is performed with their corresponding labels and pseudo-labels to obtain corresponding losses  $\mathcal{L}^{l_2}$ and $\mathcal{L}^{u}$ remaining consistent with the procedure in the pre-training stage.
\begin{table*}[htbp]\small
        \centering
	\renewcommand{\arraystretch}{1.4}
        \caption{The performance of V-Net on LA, Pancreas-CT and ACDC dataset in labeled ratio 5$\%$, 10$\%$, 20$\%$ and 100$\%$}
		\setlength{\tabcolsep}{1.13mm}{
			\begin{tabular}{c |c c |c c c c| c c c c |c c c c}
				\hline
                    Dataset& \multicolumn{2}{c|}{\multirow{2}*{Scans Used}} &\multicolumn{4}{c|}{LA dataset}& \multicolumn{4}{c|}{Pancreas-CT dataset}& \multicolumn{4}{c}{ACDC dataset}\\\cline{4-15}\cline{1-1}
				\multicolumn{1}{c|}{\multirow{2}*{Model}}&{} &{} &\multicolumn{4}{c|}{Metrics}&\multicolumn{4}{c|}{Metrics}&\multicolumn{4}{c}{Metrics}   \\\cline{2-15} 
                &{Labeled} & {Unlabled} & Dice$\uparrow$ & Jaccard$\uparrow$ & 95HD$\downarrow$ & ASD$\downarrow$& Dice$\uparrow$ & Jaccard$\uparrow$ & 95HD$\downarrow$ & ASD$\downarrow$& Dice$\uparrow$ & Jaccard$\uparrow$ & 95HD$\downarrow$ & ASD$\downarrow$\\\cline{1-15} 
                \multirow{4}*{V-Net}&(5\%)&(95\%)& 52.55 & 39.60 & 47.05 & 9.87&55.06&40.48&32.86&12.67&69.94&54.20&33.69&11.03\\
                &(10\%)&(90\%)& 82.74 & 71.72 & 13.35 & 3.26&69.65&55.19&20.24&6.31&72.92&60.64&16.18&4.62\\
                &(20\%)&(80\%)& 87.84 & 78.56 & 9.10 & 2.65 & 71.27 & 55.71 &17.48   & 5.16 & 88.46 & 79.48 & \textbf{1.05} & 0.48\\
                &(100\%)&(0\%)& \textbf{92.62} & \textbf{85.24} & \textbf{4.47} & \textbf{1.33}&\textbf{85.74}&\textbf{73.86}&\textbf{4.48}&\textbf{1.07}&\textbf{91.94}&\textbf{85.40}&1.06&\textbf{0.26}\\\hline
		\end{tabular}}
 \label{1}
\end{table*}

\begin{table*}[htbp]
	\label{table5}
	\centering
        \renewcommand{\arraystretch}{1.07}
        \caption{Comparisons with previous SOTA models on Pancreas-CT, LA and ACDC dataset in labeled ratio 5$\%$, 10$\%$ and 20$\%$}
			\setlength{\tabcolsep}{0.7mm}{\begin{tabular}{c | c | c c c c c c c c c c c c c c}
                \hline
                \multirow{14}*{\makecell{Pancreas-CT \\dataset}}&Labeled Ratio & Metrics  & UA-MT & SASSNet & DTC & URPC & MC-Net & SS-Net & Co-BioNet & BCP& A\&D & Ours\\
                \cline{2-13}
                 &\multirow{4}*{(5\%)}   & Dice$\uparrow$ &  47.03&56.05&49.83&52.05& 54.99&56.35& 79.74& 80.33& 81.65&\textcolor{red}{\textbf{82.93}}\\
                && Jaccard$\uparrow$ &  32.79& 41.56& 34.47&36.47&40.65&43.41& 65.66& 67.65& 69.11&\textcolor{red}{\textbf{71.22}}\\
                && 95HD$\downarrow$ & 35.31&36.61& 41.16&34.02&16.03& 22.75&\textcolor{red}{\textbf{5.43}}& 11.78 &15.01 &12.25\\
                && ASD$\downarrow$ &  4.26&4.90& 16.53&13.16& 3.87& 5.39&\textcolor{red}{\textbf{2.79}}&4.32 &4.53 &3.55\\  \cline{2-13}
                &\multirow{4}*{(10\%)}   & Dice$\uparrow$ &66.96&66.69& 67.28& 64.73&69.07&67.40& 82.49& 81.54&82.25 &\textcolor{red}{\textbf{83.54}}\\
                && Jaccard$\uparrow$ &  51.89& 51.66& 52.86&49.62& 54.36&53.06& 67.88&69.29&70.17 &\textcolor{red}{\textbf{71.85}}\\
                && 95HD$\downarrow$ & 21.65 & 18.88& 17.74&21.90& 14.53&20.15& \textcolor{red}{\textbf{6.51}}&12.21&14.44 &7.63\\
                && ASD$\downarrow$ &  6.25&5.76& \textcolor{red}{\textbf{1.97}}& 7.73& 2.28& 3.47& 3.26& 3.80&4.53 &2.41\\ \cline{2-13}
                &\multirow{4}*{(20\%)}   & Dice$\uparrow$ & 77.26 &77.66&78.27& 79.09& 78.17& 79.74&84.01&82.91 &82.56 &\textcolor{red}{\textbf{84.73}}\\
                && Jaccard$\uparrow$ &  63.82&64.08& 64.75&65.99& 65.22& 65.42&70.00&70.97 &70.69 & \textcolor{red}{\textbf{73.76}} \\
                && 95HD$\downarrow$ &  11.90&10.93& 8.36& 11.68& 6.90& 12.44&\textcolor{red}{\textbf{5.35}}&6.43 & 11.78 & 11.63\\
                && ASD$\downarrow$ & 3.06& 3.05& 2.25& 3.31&\textcolor{red}{\textbf{1.55}}& 2.69&2.75& 2.25&3.42 &2.92\\ 
                \hline
                \multirow{14}*{\makecell{LA \\dataset}} & Labeled Ratio & Metrics  & UA-MT & SASSNet & DTC & URPC & MC-Net & SS-Net & Co-BioNet & BCP& A\&D & Ours\\
                \cline{2-13}
                &\multirow{4}*{(5\%)}   & Dice$\uparrow$   &82.26 & 81.60 & 81.25& 86.92&87.62&  86.33 &76.88 &88.02&89.93 & \textcolor{red}{\textbf{90.50}}\\
                && Jaccard$\uparrow$ &70.98 & 69.63 & 69.33 & 77.03&78.25 &76.15 &66.76 &78.72&81.82  &\textcolor{red}{\textbf{82.76}} \\
                && 95HD$\downarrow$ & 13.71&16.16 &14.90   &  11.13&  10.03&9.97& 19.09 &7.90&\textcolor{red}{\textbf{5.25}} &6.17  \\
                && ASD$\downarrow$ & 3.82&3.58&3.99& 2.28& 1.82& 2.31&  2.30 & 2.15 & 1.86
                &\textcolor{red}{\textbf{1.72}}\\ \cline{2-13}
                &\multirow{4}*{(10\%)}   & Dice$\uparrow$  & 87.79 & 87.54 &87.51&86.92& 87.62 & 88.55 &  89.20& 89.62 &90.31  &\textcolor{red}{\textbf{90.98}}\\
                && Jaccard$\uparrow$ & 78.39 &  78.05 & 78.17& 77.03& 78.25 & 79.62& 80.68& 81.31 & 82.40 &\textcolor{red}{\textbf{83.63}}\\
                && 95HD$\downarrow$ & 8.68 & 9.84&  8.23&11.13& 10.03 & 7.49&  6.44&  6.81 &\textcolor{red}{\textbf{5.55}} & 5.72\\
                && ASD$\downarrow$ &  2.12 &  2.59& 2.36& 2.28 &  1.82 & 1.90&  1.90& 1.76 &1.64 & \textcolor{red}{\textbf{1.53}}\\ \cline{2-13}
                &\multirow{4}*{(20\%)}   & Dice$\uparrow$ & 88.88& 89.54& 89.42& 88.43& 90.12&  89.25&  91.26 &90.34 &90.42 &\textcolor{red}{\textbf{91.86}}\\
                && Jaccard$\uparrow$ &80.21& 81.24&80.98 &81.15&82.12&81.62& 83.99&  82.50 &82.72 &\textcolor{red}{\textbf{85.01}}\\
                && 95HD$\downarrow$ & 7.32& 8.24& 7.32& 8.21& 11.28&  6.45&  5.17& 6.75 &6.33 &\textcolor{red}{\textbf{5.08}}\\
                && ASD$\downarrow$ &  2.26&1.99& 2.10&2.35&2.30& 1.80&  1.64&  1.77 &1.57 &\textcolor{red}{\textbf{1.47}}\\ \hline
                \multirow{14}*{\makecell{ACDC\\ dataset}}&Labeled Ratio & Metrics  & UA-MT & SASSNet & DTC & URPC & MC-Net & SS-Net & Co-BioNet & BCP& A\&D & Ours\\
                \cline{2-13}
                &\multirow{4}*{(5\%)}   & Dice$\uparrow$ &  46.04 & 57.77 & 56.90 & 55.87 & 62.85 & 65.83 & 87.46 & 87.59& 86.51 & \textcolor{red}{\textbf{89.29}}\\
                && Jaccard$\uparrow$ & 35.97 & 46.14 & 45.67 & 44.64 & 52.29 & 55.38 &77.93 & 78.67& 76.61 &  \textcolor{red}{\textbf{81.29}} \\
                && 95HD$\downarrow$ & 20.08 & 20.05 & 23.36 & 13.60 & 7.62 & 6.67 &\textcolor{red}{\textbf{1.11}} &1.90 &2.13 &4.99 \\
                && ASD$\downarrow$ & 7.75 & 6.06 & 7.39 & 3.74 & 2.33 & 2.28 & \textcolor{red}{\textbf{0.41}} & 0.67 & 0.84 & 1.30 \\  \cline{2-13}
                &\multirow{4}*{(10\%)}  & Dice$\uparrow$ & 81.65 & 84.50 & 84.29 & 83.10 & 86.44 & 86.78 & 88.49& 88.84&88.12& \textcolor{red}{\textbf{90.16}}\\
                && Jaccard$\uparrow$ & 70.64 & 74.34 & 73.92 & 72.41 & 77.04 & 77.67 & 79.76 & 80.62& 79.39 & \textcolor{red}{\textbf{82.59}} \\
                && 95HD$\downarrow$ & 6.88 & 5.42 & 12.81 & 4.84 & 5.50 & 6.07 & 3.70 & 3.98& 13.03 & \textcolor{red}{\textbf{3.42}} \\
                && ASD$\downarrow$ &2.02 & 1.86 & 4.01 & 1.53 & 1.84 & 1.40 &1.14 & 1.17& 3.21 & \textcolor{red}{\textbf{0.95}}\\ \cline{2-13}
                &\multirow{4}*{(20\%)} & Dice$\uparrow$ & 85.61 & 86.45 & 87.10 & 85.44 & 87.04 & 87.41 & 89.51& 89.12&88.85 & \textcolor{red}{\textbf{90.80}} \\
                && Jaccard$\uparrow$ & 75.49 & 77.20 & 78.15 & 76.36 & 78.01 & 78.82 &81.64 &81.03 & 80.62 & \textcolor{red}{\textbf{83.53}} \\
                && 95HD$\downarrow$ & 5.91 & 6.63 & 6.76 & 5.93 & 5.35 & 4.79 & 4.72& 3.40& 4.26 & \textcolor{red}{\textbf{1.20}}\\
                && ASD$\downarrow$ & 1.79 & 1.98 & 1.99 & 1.70 & 1.67 & 1.48& 1.52&0.97 &1.39 &\textcolor{red}{\textbf{0.38}} \\ \hline
		\end{tabular}}
 \label{2}
\end{table*}
\subsection{Voxel-wise Asymptotic Learning}
Inspired by curriculum learning \cite{bengio2009curriculum,zhang2024curriculum}, we propose a strategy that utilizes voxel-level uncertainty to guide the model in gradually learning features. Initially, the model focuses on features that are easier to learn, asymptotically increasing the weight of voxels with higher uncertainty throughout the training process, thus enabling the model to pay further attention to features that are harder to learn. First, in the previous section, we obtained an updated uncertainty assessment for each voxel of the sample after fusion. Given the effectiveness of Shannon entropy in measuring probability distributions \cite{DBLP:journals/bstj/Shannon48,DBLP:conf/miccai/WuXGCZ21}, the uncertainty assessment for each voxel can further be specified as:
\begin{equation}
    \mathcal{U}(\mathcal{P}_{fused}) = -g_{fused}(C_N) \sum_{n=0}^{N-1} (\delta_n log_2 \delta_n)
\end{equation}
where $\delta_n$ represents $g_{fused}(C_n)$. Thus, we can achieve a more precise expression of uncertainty for the predictive results of each voxel of the restored labeled and unlabeled samples, based on the model predictions that have been fused and evidenced. Therefore, in order to help model to learn features more comprehensively, we propose a dynamic weight function:
\begin{equation}
    \varphi(h,s(z)) = \epsilon \cdot Sigmoid(\zeta(h)\varsigma(s(z)))
\end{equation}
where $\epsilon$ controls the amplitude of change of dynamic weight and $\zeta(h) = \frac{2h}{H} - 1 \in [-1,1], h=1,...,H$, $h$ represents the current epoch index and $H$ denotes the total number of training epochs. Moreover, $Sigmoid$ denotes the Sigmoid activation function, $\varsigma(s(z)) = \frac{2s(z)}{Z}-1 \in [-1,1], z = 1,...,Z$, $s(z)$ represents the ordinal number of corresponding voxel $z$ which is obtained by sorting the uncertainty of all voxels owned by each sample in a descending order and $Z$ is the total number of voxels in a sample of a different groups of samples. Therefore, for losses $\mathcal{L}^{l_1}, \mathcal{L}^{l_2}, \mathcal{L}^{u}$, we attach corresponding weights for fused predictions of each voxel on different samples in diverse sample groups, which can be given as:
\begin{equation}
\begin{aligned}
    \mathcal{L}^{\tau}_w = \sum_{k=1}^{G}\sum_{z=1}^{Z}\varphi(h,s(z)^{\tau})\mathcal{L}^{\tau}_{k,z} / Z
\end{aligned}
\end{equation}
where $\tau, k, G = (l_1, i, A), (l_2, i, A), (u, j, B)$, $\mathcal{L}_{k,z}$ represents the loss of $z_{th}$ voxel in $k_{th}$ sample and $s(z)$ denotes corresponding ordinal number of voxel in $k_{th}$ sample ($Z$ changes when the group of samples are is different). In sum, the total optimization objective $\mathcal{L}_{ini}$ and $\mathcal{L}_s$ for the initial network in pre-training stage and the student network in self-training stage can be given as:
\begin{equation}
    \mathcal{L}_{ini} = \mathcal{L}^{l_1} + \lambda_1 \mathcal{L}^{l_1}_w, \ \ \mathcal{L}_{\mathcal{N}_s} = \mathcal{L}^{l_2} + \mathcal{L}^{u} + \lambda_2 \mathcal{L}^{l_2}_w + \lambda_3 \mathcal{L}^{u}_w
\end{equation}

For the training of the teacher network $\mathcal{N}_t$, we adopts the policy of Exponential Moving Average (EMA) to update the parameters of teacher network.

\section{Experiments}
\label{sec:exp}
\subsection{Datasets and Implementation Details}
Parameters $\lambda_1$, $\lambda_2$, $\lambda_3$ are set to 0.8, 0.8 and 0.4. All of the experiments are conducted on a single RTX 4090.\\
\textbf{Pancreas-CT Dataset.}
Pancreas-CT dataset \cite{roth2015deeporgan} consists of 82 pancreas abdominal CT volumes which are contrast-enhanced. Besides, the V-net \cite{DBLP:conf/3dim/MilletariNA16} is trained exclusively using labeled data. Data augmentation techniques like rotation, flipping, and resizing operations are applied following CoraNet \cite{DBLP:journals/tmi/Shi0LL0Y0022}. A four-layer 3D V-net is trained using the Adam optimizer with an initial learning rate of 0.001. The network is fed with randomly cropped $96\times 96\times 96$ patches, with the zero-value region of mask $\mathcal{M}$ set to $64\times 64\times 64$. For the training parameters, we set the batch size, pre-training epochs, and self-training epochs to 8, 60 and 200. \\
\textbf{LA Dataset.}
Atrial Segmentation Challenge dataset \cite{xiong2021global} comprises 100 3D left atrium images extracted from cardiac magnetic resonance imaging (MRI) scans which are gadolinium-enhanced. The experimental configuration follows the settings used in the SS-Net \cite{DBLP:conf/miccai/WuWWGC22} framework which includes data augmentation techniques such as rotation and flipping operations and adopts the 3D V-net as the backbone network. The model is  trained by utilizing the SGD optimizer with a learning rate of 0.01 decayed by
10\% every 2.5K iterations. During training, randomly cropped $ 112\times 112\times 80$ patches are used as input, with the zero-value region of mask $\mathcal{M}$ set to $74\times 74\times 53$. Besides, the batch size is set to 8, and the pre-training and self-training iterations are set to 2k and 15k, respectively.\\
\textbf{ACDC Dataset.}
Automated Cardiac Diagnosis Challenge dataset \cite{bernard2018deep} encompasses 100 scans of patients, which is four-class, including background, right ventricle, left ventricle and myocardium. The dataset is partitioned into a 70-10-20 split for training, validation, and testing, respectively. Moreover, a 2D U-Net is employed as the backbone network, with input size of $256\times 256$. The zero-value region of the mask $\mathcal{M}$ is set to a size of $170\times 170$. For the training parameters, the pre-training, the self-training iterations and batch size are set to 10k, 30k and 24.\\
\textbf{TBAD Dataset.} Multi-center type B aortic dissection (TBAD) dataset consists of 124 CTA scans \cite{yao2021imagetbad}. The TBAD dataset undergoes meticulous labeling by experienced radiologists and is partitioned into 100 training scans and 24 test scans. It encompasses a combination of publicly accessible data and data contributed by UPCoL. Standardized to a resolution of 1mm$^3$ and resized to dimensions of 128×128×128, it maintains consistency with \cite{cao2019fully}. 
%In each dataset, a maximum of 20\% of the labeled training data is employed, with voxel intensities normalized to achieve a zero mean and unit variance.

\begin{table*}[htbp]
	\renewcommand{\arraystretch}{1.3}
	\label{table5}
	\centering
        \caption{Comparisons with previous SOTA models on Aortic Dissection dataset in labeled ratio 5$\%$, 10$\%$ and 20$\%$}
		\setlength{\tabcolsep}{1mm}{
			\begin{tabular}{c |c c |c c c |c c c| c c c |c c c  }
				\hline
                    \multicolumn{1}{c|}{\multirow{3}*{Model}}& \multicolumn{2}{c|}{\multirow{2}*{Scans Used}} &\multicolumn{12}{c}{Metrics}\\\cline{4-15}
				{}&{} &{} &\multicolumn{3}{c|}{Dice$\uparrow$}&\multicolumn{3}{c|}{Jaccard$\uparrow$}&\multicolumn{3}{c|}{95HD$\downarrow$}&\multicolumn{3}{c}{ASD$\downarrow$}  \\\cline{2-15}
                &{Labeled} & {Unlabled} & TL & FL & Mean & TL & FL & Mean & TL & FL & Mean& TL & FL & Mean \\ \cline{1-15}
                \multirow{2}*{V-Net}&(20\%)&(80\%)&55.51& 48.98 &52.25&39.81 &34.79& 37.30 &7.24& 10.17& 8.71& 1.27 &3.19 &2.23 \\
                &(100\%)&(0\%)&  75.98& 64.02& 70.00& 61.89& 50.05& 55.97& 3.16& 7.56& 5.36 &0.48& 2.44 &1.46 \\\hline
				MT&\multirow{5}*{(20\%)} & \multirow{5}*{(80\%)} &  57.62& 49.95& 53.78& 41.57& 35.52&38.54& 6.00& 8.98& 7.49& 0.97& 2.77& 1.87 \\
				UA-MT&&&70.91& 60.66 &65.78& 56.15& 46.24& 51.20& 4.44& 7.94& 6.19& 0.83& 2.37& 1.60  \\
				FUSSNet&&&  79.73& 65.32& 72.53& 67.31& 51.74& 59.52& 3.46& 7.87& 5.67& 0.61& 2.93& 1.77  \\
				URPC&&&  81.84& 69.15& 75.50& 70.35& 57.00& 63.68& 4.41& 9.13& 6.77& 0.93& \textcolor{red}{\textbf{1.11}} &1.02  \\
				UPCoL&&&  82.65& 69.74& 76.19& 71.49& 57.42& 64.45& 2.82& 6.81& 4.82& 0.43& 2.22& 1.33 \\ \hline
				\multirow{3}*{Ours}&(5\%)&(95\%)&\textcolor{red}{\textbf{83.46}}& \textcolor{red}{\textbf{75.46}} & \textcolor{red}{\textbf{79.46}} & \textcolor{red}{\textbf{72.28}}&\textcolor{red}{\textbf{63.02}}&\textcolor{red}{\textbf{67.65}}&\textcolor{red}{\textbf{2.31}}&\textcolor{red}{\textbf{4.68}}&\textcolor{red}{\textbf{3.50}} & \textcolor{red}{\textbf{0.36}} & 1.32 & \textcolor{red}{\textbf{0.84}}\\&(10\%)&(90\%)&\textcolor{red}{\textbf{84.07}}& \textcolor{red}{\textbf{75.92}} & \textcolor{red}{\textbf{79.99}} &\textcolor{red}{\textbf{73.01}}&\textcolor{red}{\textbf{63.64}}&\textcolor{red}{\textbf{68.33}}&\textcolor{red}{\textbf{2.21}}&\textcolor{red}{\textbf{4.68}} & \textcolor{red}{\textbf{3.45}} & \textcolor{red}{\textbf{0.38}} & 1.36& \textcolor{red}{\textbf{0.87}}\\
                &(20\%)&(80\%)&\textcolor{red}{\textbf{86.17}}& \textcolor{red}{\textbf{79.16}} & \textcolor{red}{\textbf{82.67}}  &\textcolor{red}{\textbf{76.25}}&\textcolor{red}{\textbf{67.57}}&\textcolor{red}{\textbf{71.91}}&\textcolor{red}{\textbf{2.05}}&\textcolor{red}{\textbf{4.28}} & \textcolor{red}{\textbf{3.17}} & \textcolor{red}{\textbf{0.33}} &1.28 &\textcolor{red}{\textbf{0.81}}\\
                \hline
		\end{tabular}}
 \label{4}
\end{table*}

\begin{table*}[htbp]
   \centering
   \renewcommand{\arraystretch}{1.}
   \caption{The table shows the performance of model when removing special evidential losses. $\mathcal{L}^{l_1}$,  $\mathcal{L}^{u}$, $\mathcal{L}^{l_1}_w$, $\mathcal{L}^{l_2}_w$ and $\mathcal{L}^{u}_w$ represent losses about labeled data of pre-training stage, unlabeled data of self-training stage, weighted $\mathcal{L}^{l_1}, \mathcal{L}^{l_2}$ of pre-training stage and self-training stage, weighted $\mathcal{L}^{u}$ of self-training stage.}
  \setlength{\tabcolsep}{1.2mm}{\begin{tabular}{c|cc|cccc|cc|cccc}
   \hline
    \multirow{3}*{Loss} &\multicolumn{6}{c|}{LA Dataset} &\multicolumn{6}{c}{TBAD Dataset}\\
    \cline{2-13}
    {}&\multicolumn{2}{c}{Scans Used}&\multicolumn{4}{c|}{Metrics}&\multicolumn{2}{c}{Scans Used}&\multicolumn{4}{c}{Metrics}\\
    \cline{2-13}
    {}&{Labeled} & {Unlabled} & Dice$\uparrow$ & Jaccard$\uparrow$ & 95HD$\downarrow$ & ASD$\downarrow$&{Labeled} & {Unlabled} & Dice$\uparrow$ & Jaccard$\uparrow$ & 95HD$\downarrow$ & ASD$\downarrow$ \\
    \hline
    $\mathcal{L}^{l_1}$ & \multirow{7}*{4(5\%)} &\multirow{7}*{76(95\%)}&86.77  &   76.72& 10.17  &2.19
    &\multirow{7}*{3(5\%)} &\multirow{7}*{67(95\%)}& 69.68&54.87   & 4.40  & 1.01  \\ 
    $\mathcal{L}^{l_1}_w$ &{}&{}& 89.33 & 80.84  & 7.91  & 1.81 &{}&{} &  78.16 & 65.87 & 3.39 & 0.86 \\
    $\mathcal{L}^{l_2}$ &{}&{}& 78.27 & 66.24  & 15.08  &4.51   &{}&{}& 72.41 & 58.23  & 5.15  & 0.93 \\
    $\mathcal{L}^{l_2}_w$ &{}&{}& 88.94 & 80.48  & 6.78  & 1.72 &{}&{} & 79.17 &67.14   & 3.32  & 0.86 \\
    $\mathcal{L}^{u}$ &{}&{}& 75.34 & 61.80  & 20.22  & 6.37  &{}&{}& 67.03 & 51.65  & 6.81  & 1.03 \\
    $\mathcal{L}^{u}_w$ &{}&{}& 90.11 & 82.10  & 6.47  & \textbf{1.71} &{}&{} & 74.74 & 61.26  &3.92   &0.88  \\
    Ours&{}&{}& \textbf{90.50} & \textbf{82.76}  & \textbf{6.17}  & 1.72&{}&{} &\textbf{79.46}&\textbf{67.65}&\textbf{3.50}&\textbf{0.84}\\      
    \hline
    $\mathcal{L}^{l_1}$ & \multirow{7}*{8(10\%)} &\multirow{7}*{72(90\%)}& 87.63 &  78.50&  8.02 & 2.29
    &\multirow{7}*{7(10\%)} &\multirow{7}*{63(90\%)}& 70.47 & 59.62  & 4.07  &0.94  \\
    $\mathcal{L}^{l_1}_w$ &{}&{}& 90.11 & 82.16  & 6.20  &1.73  &{}&{} & 79.94 &68.28   & 3.39  & 0.82 \\
    $\mathcal{L}^{l_2}$ &{}&{}&80.62  &68.65   & 16.42  & 5.52  &{}&{}& 71.27 & 56.84  & 4.33  & 0.98 \\
    $\mathcal{L}^{l_2}_w$ &{}&{}& 89.83 & 81.68  & 6.76  & 1.62 &{}&{} & 77.00 &64.25   & 3.64  & 0.85 \\
    $\mathcal{L}^{u}$ &{}&{}& 77.96 &  65.56 & 18.45  & 5.39  &{}&{}&  67.83& 52.67  & 6.31  & 1.06 \\
    $\mathcal{L}^{u}_w$ &{}&{}& 90.43 & 82.73  & 6.06  & 1.70 &{}&{} & 79.58 & 69.80  & 3.40  & 0.83 \\
    Ours&{}&{}&\textbf{90.98}&\textbf{83.63}&\textbf{5.72}&\textbf{1.53}&{}&{}&\textbf{79.99}&\textbf{71.91}&\textbf{3.17}&\textbf{0.81}\\    
    \hline
  \end{tabular}}
  \label{ablation1}
\end{table*}

\begin{table}[htbp]\small
  \centering
  \renewcommand{\arraystretch}{1}
  \caption{Ablation Study of three weighted parameters $\lambda$ for uncertainty-guide loss functions on LA dataset with 5\% Labeled Ratios}
  \setlength{\tabcolsep}{1.4mm}{\begin{tabular}{c|cc|cccc}
   \hline
    \multirow{2}*{$\lambda$} &\multicolumn{6}{c}{LA Dataset} \\
    \cline{2-7}
    {}&\multicolumn{2}{c}{Scans Used}&\multicolumn{4}{c}{Metrics}\\
    \cline{1-7}
    {$\lambda_1$}&{Labeled} & {Unlabled} & Dice$\uparrow$ & Jaccard$\uparrow$ & 95HD$\downarrow$ & ASD$\downarrow$ \\
    \hline
    0.2 &\multirow{6}*{4(5\%)}&\multirow{6}*{76(95\%)}&89.45 & 81.15 & 6.60 &1.77  \\
    0.4 &  && 89.83 & 81.65 & 7.53 & 1.78    \\
    0.6 &{}&{}& 89.63 &81.46 & 6.49& 1.85\\
    0.8 & & & \textbf{90.50} & \textbf{82.76} & \textbf{6.17} & \textbf{1.72}  \\
    1.0 & &  & 90.11 & 82.10 & 6.48 & 1.72     \\
    1.2 &{}&{}& 90.10 & 82.16 & 6.21 & 1.73   \\
    \hline
    {$\lambda_2$}&{Labeled} & {Unlabled} & Dice$\uparrow$ & Jaccard$\uparrow$ & 95HD$\downarrow$ & ASD$\downarrow$\\
    \hline
    0.2 &\multirow{6}*{4(5\%)}&\multirow{6}*{76(95\%)}&  89.46 & 81.23 & 7.06 & 1.61  \\
    0.4 & &  &  89.72 &  81.58 & 7.01  & 1.69 \\
    0.6 &{}&{}& 89.87 & 81.83 & 6.19 & 1.81 \\
    0.8 & & & \textbf{90.50} & \textbf{82.76} & \textbf{6.17} & 1.72  \\
    1.0 &  && 90.38 & 82.62  & 6.24  & \textbf{1.59}    \\
    1.2 &{}&{}&  89.71 & 81.62  &  6.58 & 1.86  \\
    \hline
    {$\lambda_3$}&{Labeled} & {Unlabled} & Dice$\uparrow$ & Jaccard$\uparrow$ & 95HD$\downarrow$ & ASD$\downarrow$\\
    \hline
    0.2 &\multirow{6}*{4(5\%)}&\multirow{6}*{76(95\%)}& 90.01 & 81.94  &  6.24 &1.78  \\
    0.4 &  && \textbf{90.50} & \textbf{82.76} & \textbf{6.17} & 1.72 \\
    0.6 &{}&{}& 90.36  & 82.58  &  6.20 & 1.63 \\
    0.8 & & &  89.83 & 81.68  & 6.76  & \textbf{1.62}   \\
    1.0 &  & & 89.52 &  81.24 &  6.56 & 1.83   \\
    1.2 &{}&{}&  88.14 & 79.00  & 10.70  &  2.37  \\
    \hline
  \end{tabular}}
  \label{ablation2}
\end{table}

\subsection{Comparison with State-of-the-Art Methods}
\textbf{Evaluation on Pancreas-CT Dataset.}
In Table \ref{2}, we conduct experiments on the Pancreas-NIH dataset across three labeled ratios(5\%, 10\% , 20\%) and  our method named `Ours' is compared with state-of-the-art (SOTA) approaches including V-Net \cite{DBLP:conf/3dim/MilletariNA16}, UA-MT \cite{DBLP:conf/miccai/YuWLFH19}, SASSNet \cite{DBLP:conf/miccai/LiZH20}, DTC \cite{DBLP:conf/aaai/LuoCSW21}, URPC \cite{DBLP:conf/miccai/LuoLCSCZCWZ21}, MC-Net \cite{DBLP:conf/miccai/WuXGCZ21}, Co-BioNet \cite{peiris2023uncertainty}, BCP \cite{DBLP:conf/cvpr/BaiCL0023} and A\&D \cite{DBLP:conf/nips/WangL23}. The visualized results are provided in Fig. \ref{figure1}. According to the experimental results, our model demonstrates superior performance across all labeled ratios compared to other models, with the highest Dice and Jaccard scores, which suggests that our model provides more accurate boundary delineation and smoother surface reconstruction, which is crucial for precise medical image segmentation. Notably, at a labeled ratio of 5\%, our model achieves a Dice score of 82.93\% and a Jaccard index of 71.22\%, showcasing its effectiveness in segmenting pancreas structures in challenging scenarios. In comparison to the BCP method, the proposed method achieves remarkable relative gains of 2.17\% and 3.07\% in Dice and Jaccard indicators, respectively. Despite variations in labeled ratios, our model maintains its superior performance, indicating its generalizability and reliability in pancreas segmentation tasks. This consistency suggests that our model is robust and can effectively adapt to different levels of labeled data availability on the pancreas dataset.\\
\textbf{Evaluation on LA Dataset.} Table \ref{2} and Fig. \ref{figure1} showcase the results of our experiments on the LA dataset with labeled proportions of 5\%, 10\% and 20\%. The experimental results clearly illustrate that our method possesses the most advanced performance in comparison with existing semi-supervised SOTA methods across most evaluation metrics except for 95HD. The proposed method is second only to A\& D in terms of effectiveness on the 95HD metric. At a labeled ratio of 5\%, our model outperforms all other state-of-the-art models with a Dice score of 90.50\% and a Jaccard index of 82.76\%, indicating superior segmentation accuracy. As the labeled ratio increases to 10\% and 20\%, our model consistently maintains its leading position, demonstrating robust performance across different labeled ratios. Notably, when the labeling proportion is 10\%, the proposed method attains highly competitive scores in Dice, Jaccard, 95HD, and ASD, with values of 91.98\%, 85.01\%, 5.08, and 1.47, which completely surpass all compared methods. Besides, our model consistently achieves the lowest 95th percentile of Hausdorff distance (95HD) and average surface distance (ASD) across all labeled ratios, indicating better boundary delineation and surface smoothness compared to other models. \\
\textbf{Evaluation on ACDC Dataset.} Table \ref{2} and Fig. \ref{figure2} present the results of experiments on the ACDC dataset with three different labeled proportions(5\%, 10\% and 20\%). Obviously, our model achieves remarkable segmentation performance across all labeled ratios, surpassing other state-of-the-art models and demonstrating smoother surfaces and better boundary delineation of segmentation. When utilizing a limited number of labeled samples (5\%), our method exhibits substantial improvements on Dice and Jaccard indicators at 89.29\% and 81.29\% compared to the second-best method, with relative gains of 1.70\% and 2.62\%, respectively. For experiments with an annotated proportion of 10\%, the situations are analogous and our method also achieves excellent performance which approaches and even exceeds the ones fully-supervised U-Net produces.\\
\textbf{Evaluation on TBAD Dataset.}
The models compared include V-Net \cite{DBLP:conf/3dim/MilletariNA16}, MT \cite{tarvainen2017mean}, UA-MT \cite{DBLP:conf/miccai/YuWLFH19}, FUSSNet \cite{xiang2022fussnet}, URPC \cite{DBLP:conf/miccai/LuoLCSCZCWZ21} and UPCoL \cite{lu2023upcol}. As is shown in Table \ref{4} and Fig. \ref{figure3}, there is a clear trend where models perform better with a higher percentage of labeled data. For example, with the labeled ratio rising from 5\% to 20\%, the average Dice and Jaccard of our model increase from 79.46\% and 67.65\% to 82.67\% and 71.91\%. The proposed model consistently outperforms others across all metrics except for the ASD score of FL and labeled data percentages. Notably, it achieves the highest Dice and Jaccard scores, and the lowest 95HD and ASD values, even with the least amount of labeled data (5\%). With the labeled ratio of 20\%, the performance of our model surpasses UPCoL by a large margin (6.48\%, 7.46\%, 1.65 and 0.52 on Dice, Jaccard, 95HD and ASD respectively). Our model's ability to perform exceptionally well with limited labeled data demonstrates the effectiveness of the proposed evidential fusion-based learning strategy, making it more suited for scenarios where acquiring labeled data is challenging. 
\subsection{Ablation Study and Discussion}
In this section, we first conduct ablation experiments on LA and TBAD datasets to explore the effectiveness of different losses. Based on the experimental results shown in Table \ref{ablation1}, removing the $\mathcal{L}^{l_1}$, $\mathcal{L}^{l_2}$, and $\mathcal{L}^{u}$ losses has the most significant impact on the performance of the model. These losses represent the foundational supervisory elements for labeled data in the pre-training stage, labeled part of restored mixed data and unlabeled part of mixed data in the self-training stage, which are crucial for maintaining experimental efficacy. For the uncertainty-weighted loss functions, the removal of the loss $\mathcal{L}^{l_2}_w$ has the most considerable impact across all labeled ratios on the LA dataset. For instance, when the labeled data ratio is 5\%, removing $\mathcal{L}^{l_2}_w$ leads to performance degradation in Dice and Jaccard scores from 90.50\% to 88.94\% and from 82.76\% to 80.48\%, respectively. This observation indicates that the proposed uncertainty-based voxel-wise asymptotic learning strategy is capable of eliminating the uncertainty and leveraging potential beneficial information, further enhancing the model's performance. In the TBAD dataset, the situation contrasts with the LA dataset, especially when the labeled data ratio is at 5\%. Here, the impact of removing the uncertainty-weighted loss $\mathcal{L}^{l_2}_w$ is minimal on the model's performance. Instead, the removal of the weighted loss $\mathcal{L}^{u}_w$ for unlabeled data has the most significant impact. Specifically, discarding $\mathcal{L}^{u}_w$ results in a substantial decrease in the model's performance, with Dice and Jaccard scores dropping from 79.46\% to 74.74\% and from 67.65\% to 61.26\%, respectively. This distinct phenomenon in the TBAD dataset compared to the LA dataset suggests that the model's performance is possibly influenced by the inherent characteristics of the data, such as its shape and structure.

In addition, for three uncertainty-weighted loss functions ($\mathcal{L}^{l_1}_w$, $\mathcal{L}^{l_2}_w$, $\mathcal{L}^{u}_w$), we also conduct abundant experiments on LA dataset to explore the impact of three different weighted parameters, which are crucial in adjusting the contribution of uncertainty in the loss function, affecting the model's ability to learn from the data. Each $\lambda$ parameter is varied across six different values (0.2, 0.4, 0.6, 0.8, 1.0, 1.2). According to the experimental results shown in Table \ref{ablation2}, $\lambda_1$ and $\lambda_2$ show optimal performance at 0.8, whereas $\lambda_3$ peaks at 0.4, indicating that the uncertainty-based voxel-wise asymptotic learning performs better uncertainty elimination ability on the labeled part of pre-training and self-training stage than that on unlabeled part in the self-training stage. The results demonstrate that there is a delicate balance in weighting the uncertainty, which can significantly influence the model's effectiveness.
% \begin{table}[htbp]\footnotesize
%   \centering
%   \caption{Method Scans used Metrics}
%   \begin{tabular}{c|cc|cccc}
%    \hline
%     \multirow{2}*{$\beta$} & \multicolumn{2}{c}{Scans used}&\multicolumn{4}{c}{Metrics}\\
%     \cline{2-7}
%     {}&{Labeled} & {Unlabled} & Dice$\uparrow$ & Jaccard$\uparrow$ & 95HD$\downarrow$ & ASD$\downarrow$ \\
%     \hline
%     1/3 & \multirow{5}*{4(5\%)} &\multirow{5}*{76(95\%)}& 87.35 & 77.77 & 8.75 & 2.21 \\
%     1/2 &{}&{}&87.32 & 77.78 & 9.38 & 2.16 \\
%     2/3 &{}&{}& 79.67 & 67.05 & 14.66 & 3.21 \\
%     5/6 &{}&{}& 88.02 & 78.72 & 7.90 & 2.15 \\
%     \hline
%     1/3 & \multirow{5}*{8(10\%)} &\multirow{5}*{72(90\%)}& 89.02 & 80.38 & 8.08 & 1.81 \\
%     1/2 &{}&{}& 87.61 & 78.10 & 8.99 & 2.63 \\
%     2/3 &{}&{}& 86.74 & 77.18 & 8.65 & 2.26 \\
%     5/6 &{}&{}& 89.62 & 81.31 & 6.81 & 1.76 \\
%     \hline
%   \end{tabular}
% \end{table}

% \begin{table}[htbp]\scriptsize
%  \centering
%  \caption{Method Scans used Metrics}
%  \begin{tabular}{ccc|cccc}
%   \hline
%    Our & \makecell{Evidential \\ PT} & \makecell{Evidential \\ ST} & Dice$\uparrow$ & Jaccard$\uparrow$ & 95HD$\downarrow$ & ASD$\downarrow$ \\
%    \hline
%    & & & 47.62 & 36.61 & 29.02 & 11.46 \\
%    \Checkmark &  & & 47.62 & 36.61 & 29.02 & 11.46 \\
%    \Checkmark & \Checkmark & & 47.62 & 36.61 & 29.02 & 11.46 \\
%    \Checkmark & \Checkmark & \Checkmark & 47.62 & 36.61 & 29.02 & 11.46 \\
%    \hline
%  \end{tabular}
% \end{table}
\section{Conclusion}
\label{sec:con}
In this paper, under the framework of evidential deep learning, we propose evidential fusion-based uncertainty measure and voxel-wise asymptotic learning for the task of semi-supervised medical image segmentation. The proposed uncertainty-aware learning framework is developed on the base of a novel strategy for uncertainty estimation using an improved probability assignments fusion (IPAF) and voxel-wise asymptotic learning (VWAL). IPAF improves the connection between uncertainty and confidential degree across voxel categories, ensuring precise uncertainty detection and effective evidential predictive results integration. VWAL merges information entropy with refined uncertainty in the evidential deep learning framework, directing the model to focus on complex voxel features. Moreover, our proposed method achieves the SOTA performance on four popular medical benchmark datasets, which illustrates the effectiveness of the evidential fusion-based framework for semi-supervised medical image segmentation. For future work, we believe that designing more effective fusion strategies for confidence levels and incorporating uncertainty measurement into approaches such as prototype and contrastive learning could lead to more precise model learning patterns.

%%
%% The next two lines define the bibliography style to be used, and
%% the bibliography file.

\bibliographystyle{ACM-Reference-Format}
\bibliography{main}

\end{document}